\documentclass[a4paper, 10pt, conference]{ieeeconf} 
                                      
\usepackage{FG2025}

\FGfinalcopy 

\IEEEoverridecommandlockouts                              
\usepackage{graphicx}
\usepackage{amsmath}
\usepackage{amssymb}
\usepackage{booktabs}
\usepackage{hyperref}
\usepackage{colortbl}

\newcommand{\name}{\textsc{OpenFace 3.0}}
\definecolor{lightblue}{RGB}{173, 216, 230}

\def\FGPaperID{256} 

\title{\LARGE \bf
{\name}: A Lightweight Multitask System\protect\\for Comprehensive Facial Behavior Analysis
}

\author{\parbox{16cm}{\centering
    {\large Jiewen Hu$^1$, Leena Mathur$^1$, Paul Pu Liang$^2$, Louis-Philippe Morency$^1$}\\
    {\normalsize
    $^1$ Carnegie Mellon University, 
    $^2$ Massachusetts Institute of Technology}\\\small\url{https://github.com/CMU-MultiComp-Lab/OpenFace-3.0}}
}

\begin{document}

\ifFGfinal
\thispagestyle{empty}
\pagestyle{empty}
\else
\pagestyle{plain}
\fi
\maketitle

\begin{abstract}
In recent years, there has been increasing interest in automatic facial behavior analysis systems from computing communities such as vision, multimodal interaction, robotics, and affective computing. Building upon the widespread utility of prior open-source facial analysis systems, we introduce \name, an open-source toolkit capable of facial landmark detection, facial action unit detection, eye-gaze estimation, and facial emotion recognition. \name\  contributes a lightweight unified model for facial analysis, trained with a multi-task architecture across diverse populations, head poses, lighting conditions, video resolutions, and facial analysis tasks. By leveraging the benefits of parameter sharing through a unified model and training paradigm, \name\ exhibits  improvements in prediction performance, inference speed, and memory efficiency over similar toolkits and rivals state-of-the-art models. \name\ can be installed and run with a single line of code and operate in real-time without specialized hardware. 
\name\ code for training models and running the system is freely available for research purposes and supports contributions from the community.
\end{abstract}

\section{Introduction} 
\vspace{-1mm}

Understanding communicative signals from faces is a critical ability driving human face-to-face social interactions \cite{haugh2009face}. A slight shift in a person's eye gaze or tilt of the head, for example, are subtle facial behaviors that can substantially influence the social meaning being conveyed and interpreted during  interactions \cite{mathur2024advancing}. Among the computer vision community, there has been a growing interest in designing systems that can use fine-grained facial behaviors to interpret human affective, cognitive, and social signals \cite{8373812, amos2016openface}. Facial behavior analysis has been a core component in
human-centered artificial intelligence (AI) systems to support human well-being, with applications in \textit{healthcare} such as detecting depression \cite{ringeval2019avec} and post-traumatic stress \cite{tavabi2020computer}, applications in \textit{education} such as robots teaching students while adapting to facial cues \cite{park2019model}, and additional use cases in automotive \cite{liu2021empathetic}, manufacturing, and service industries \cite{eyben2010emotion, hu2020robust}.

\begin{figure*}[ht]
  \centering
  \includegraphics[width=0.95\textwidth]{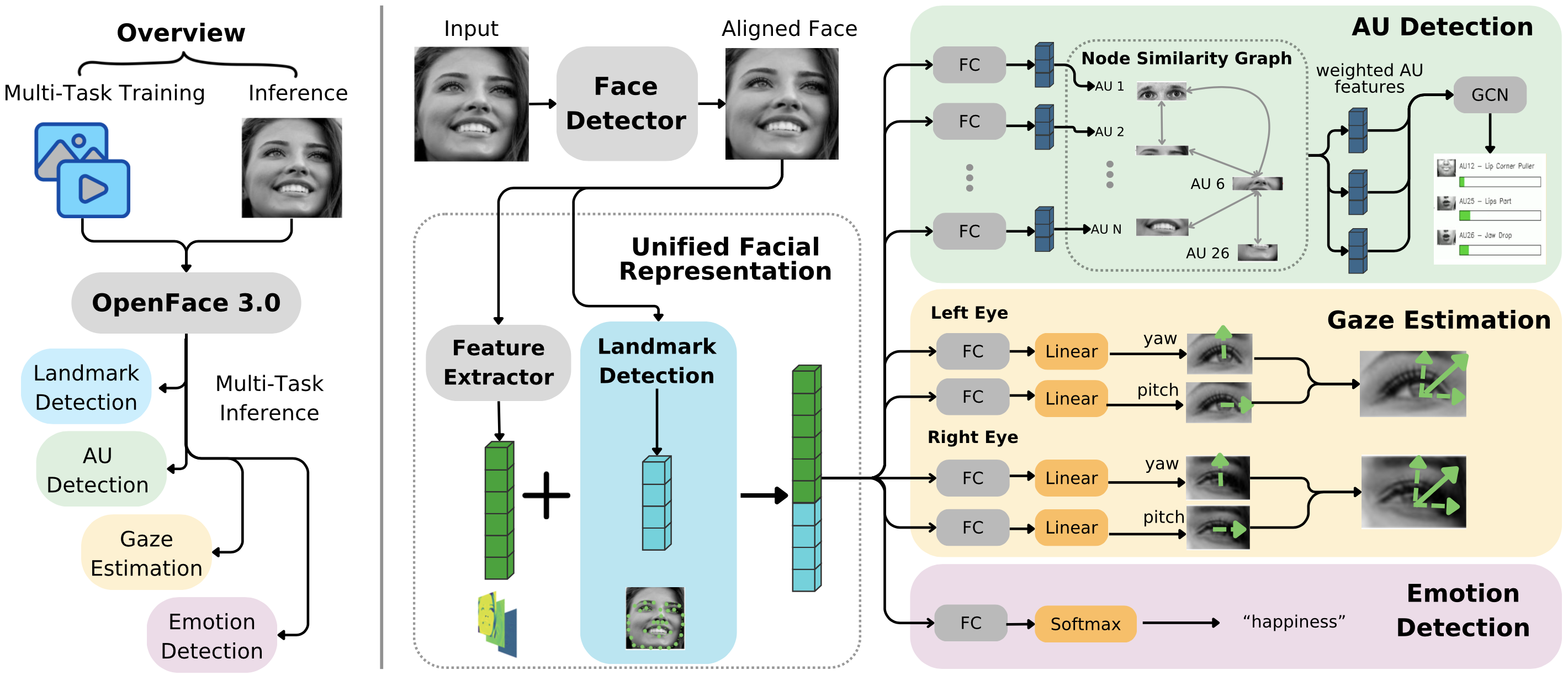}
  \caption{Visualization of the \name\ system, a lightweight multi-task modeling approach, trained for efficient facial landmark detection, AU detection, gaze estimation, and emotion recognition. During inference, the system first utilizes a face detector and feature extractor to obtain a unified facial representation that combines contextual facial information with precise facial landmark data. This unified representation is, then, used as input to three separate modules: (1) The \textit{AU detection module} constructs a similarity graph among action unit feature representations and uses a GCN to update each action unit vector. (2) The \textit{Gaze Estimation module} employs separate fully-connected (FC) layers to analyze yaw and pitch angles for each eye independently. (3) The \textit{Emotion Recognition module} uses an FC layer to classify emotions.}
  \label{fig:MTL}
\end{figure*}

This paper introduces \textbf{\name}, a high-performing open-source toolkit capable of jointly performing facial landmark detection \cite{wu2019facial}, facial action unit (AU) detection \cite{ekman1978facial},  eye gaze estimation \cite{emery2000eyes}, and facial emotion recognition  \cite{frith2009role}. These facial cues play an important role in conveying and shaping information during  interactions. \name\ has been designed to help researchers and developers easily integrate these facial signals into their own projects. 

To achieve this, \name\ contributes a new lightweight unified model for facial analysis, trained with a multi-task architecture across diverse populations, head poses, lighting conditions, video resolutions, and output multiple facial analysis tasks. \name\ achieves prediction performances that either exceed or rival larger state-of-the-art (SOTA) models that were specialized for each individual task. \name\ also performs a wider variety of tasks, in comparison to prior facial analysis toolkits \cite{amos2016openface, 8373812, Chang_2024_WACV, cheong2023py, bishay2022affdex, Jocher_Ultralytics_YOLO_2023}, and exhibits significant improvements in task performance, efficiency, inference speed, and memory usage. For example, \name\ is more accurate, faster, and more memory-efficient than \textsc{OpenFace 2.0} \cite{8373812} and achieves improved accuracy on non-frontal faces, a crucial capability for facial behavior analysis in-the-wild. 

Users can install and run \name\ with a single line of code to extract diverse facial cues in real-time and build upon an intuitive Python API to flexibly adapt our system to their needs. Our usability studies demonstrated that \name\ is user-friendly, allowing even users with limited technical expertise to leverage its capabilities. The wide variety of facial analysis tasks performed by \name, coupled with the performance and efficiency improvements of this open-source toolkit, will enable future researchers and developers across fields to create new and useful applications of facial behavior analysis. 


\vspace{-1mm}
\section{Related Work} 
\vspace{-1mm}

\subsection{Task-Specific Facial Behavior Analysis} Landmark detection approaches have evolved from traditional statistical models to deep learning approaches, such as convolutional neural networks (CNNs) \cite{wu2017facial} and heatmap-based methods to handle variations in expressions and lighting more effectively than prior models \cite{lan2022hihaccuratefacealignment, kar2024fiducial}. For AU detection, CNNs replaced earlier handcrafted feature methods, with the latest advances incorporating Graph Neural Networks (GNNs) to model relationships among facial regions \cite{baltruvsaitis2015cross, baltrusaitis2013constrained, yuce2015discriminant, corneanu2018deep, gudi2015deep, liu2014facial, zhou2017pose, wang2024multi, liu2024norface, yuan2024auformer}. Recent improvements in CNN and Vision Transformer (ViT) architectures have advanced gaze estimation by moving beyond traditional geometric models to methods that capture features from each eye, using multi-stream processing and dilated convolutions to handle varied pose and lighting \cite{tan2002appearance, zhang2015appearance, fischer2018rt, krafka2016eye, zhang2017s, cheng2022gaze}. Similarly, facial emotion recognition has progressed from using classical machine learning models \cite{sikka2013multiple} to developing CNNs and ViTs to process subtle expressions \cite{chirra2021virtual, raghav2021ensemble, roy2016local, bhattacharjee2019pattern, liu2021end, xie2020facial, xue2021transfer}. For comprehensive reviews, we refer readers to survey papers on detecting facial landmarks \cite{wu2019facial}, AUs \cite{zhi2020comprehensive},  gaze \cite{kar2017review}, and  emotion \cite{canal2022survey}. 
\subsection{Foundation Models for Computer Vision} 
In recent years, vision researchers have focused on training foundation models \cite{bommasani2021opportunities} on vast amounts of unlabeled data before transferring models to specific downstream tasks via fine-tuning. 
Originally popularized in the natural language processing literature~\cite{devlin2019bert,radford2019language}, it has become common practice to pre-train ViTs~\cite{dosovitskiy2020image} or multimodal architectures~\cite{li2019visualbert,lu2019vilbert,Su2020VLBERT,liang2022high, tsai2019multimodal} on vast amounts of unlabeled images using multitask and self-supervised objectives~\cite{caruana1997multitask,dosovitskiy2020image}. In recent years, transformer models have been leveraged as backbones in multitask learning for facial behavior analysis, due to strong feature extraction capabilities~\cite{qin2023swinface, narayan2024facexformer, zhu2024cross}.
Our work aligns with this direction, using multi-task learning to yield a more generalizable face representation that can be transferred to a variety of downstream facial analysis tasks, leading to strong task-specific performance, while remaining efficient through parameter and dataset sharing.

\subsection{Facial Analysis Toolkits}
Facial analysis toolkits allow researchers to study facial features in data from their respective domains. \textsc{OpenFace} \cite{baltruvsaitis2016openface} and \textsc{OpenFace 2.0} \cite{8373812} have been among the most widely-used open-source facial analysis systems, providing robust landmark and AU recognition capabilities. Similarly, toolkits such as LibreFace \cite{Chang_2024_WACV}, Py-Feat \cite{cheong2023py}, AFFDEX \cite{bishay2022affdex} and YOLO \cite{Jocher_Ultralytics_YOLO_2023} have gained popularity in recent years. Most prior toolkits have specialized models for a \textit{small} number of correlated tasks (e.g., AU and landmark detection). Through \name\ we contribute a new unified multi-task model trained with diverse facial analysis tasks, enabling our toolkit to effectively perform a wider range of downstream facial analysis, compared to capabilities of prior toolkits.


\vspace{-1mm}
\section{The \name\ System} 
\vspace{-1mm}




\name\ employs a multitask framework to enhance efficiency and cross-task learning by training shared facial representations across diverse tasks. Figure \ref{fig:MTL} visualizes an overview of the \name\ architecture. The model first utilizes an efficient backbone to extract unified facial features that include contextual facial information, as well as precise facial landmarks (Section \ref{sec:f1}). These unified feature representations are, then, used for diverse downstream facial analysis tasks, including AU detection, gaze estimation, and emotion recognition (Section \ref{sec:f2}). In this section, we discuss our model, as well as the training details (Sections \ref{sec:f3}, \ref{sec:f4}) and system interface (Section \ref{f5}). 


\subsection{Landmark Detection \& Unified Facial Representation}
\label{sec:f1}

The first stage includes both a \textit{facial landmark detector} to detect and align facial images and a \textit{feature extractor} that processes these aligned facial images. The resulting landmarks are combined with contextual features, creating a unified input representation with both precise localization from the landmarks and rich contextual information.

Input facial images are first passed through RetinaFace \cite{deng2020retinaface}, an open-source tool for face detection and alignment that predicts face confidence scores and bounding boxes. We use RetinaFace \texttt{MobileNet-0.25}  \cite{howard2017mobilenets}, which achieves real-time face detection and alignment, even on CPUs. 

Once the face is detected and aligned, we extract facial landmarks, which provides important, localized spatial information. For example, landmarks capture subtle facial movements, communicating different social behaviors and emotions \cite{zheng2023poster}. Similarly landmarks around the eyes partially encode information related to eye movement and gaze. We train a dedicated landmark detection module  to extract accurate landmark information. The module includes four stacked Hourglass (HG) networks as the backbone \cite{xu2021graph, yang2017stacked}, takes the aligned face as input, and outputs landmark coordinates. Each HG generates $N$ heatmaps, where $N$ is the number of pre-defined facial landmarks. The normalized heatmap can be viewed as the probability distribution over the predicted facial landmarks. The final landmark coordinates are decoded from the heatmaps using a soft-Argmax operation, ensuring smooth and differentiable predictions. (Training details for our landmark detection module are in Sections \ref{sec:f3} and \ref{sec:f4}). 

Beyond landmarks, we extract contextual features using EfficientNet \cite{tan2020efficientnetrethinkingmodelscaling} backbone for facial feature extraction, pretrained on the VGGFace2 dataset \cite{cao2018vggface2} for face recognition. We use the output of this model's final hidden layer as the contextual facial features for remaining tasks.

To fully leverage localized, spatial information from landmarks and contextual information from facial features, we concatenate both the landmarks and facial features to form a unified facial feature representation.

\subsection{Downstream Tasks: Action Unit Detection, Gaze Estimation, Emotion Recognition}
\label{sec:f2}

Our model addresses three downstream tasks that take the unified facial feature representation as input: AU detection, gaze estimation, and emotion recognition.

\textbf{AU detection}: Recognizing that relationships among AUs will shift as facial muscles move during expressions, we develop a dynamic graph-based approach inspired by ME-GraphAU \cite{Luo_2022} to model relationships among AUs. Instead of learning a static graphical relation among AUs from training data, we train a dynamic graph approach that reconstructs a \textit{new} graph for each input face based on cosine similarity metrics among AU features. 

For $N$ action units, the unified representation is passed to $N$ independent fully-connected (FC) layers to extract vector representations for each individual AU. Our model, then, constructs a similarity graph by taking $N$ AU vector representations as nodes and assigning weights to edges based on cosine similarity between AU vector representations. A Graph Convolutional Network (GCN) layer then updates each AU vector, considering both the individual characteristics of an AU and other correlated AU features.


\textbf{Gaze estimation}: We implement four independent FC layers to analyze yaw and pitch angles separately for each eye. Specifically, we separate computations for left eye yaw, left eye pitch, right eye yaw, and right eye pitch. Each  layer takes the unified representation as input and outputs the gaze vector for the corresponding angle. Informed by L2CS-NET \cite{abdelrahman2023l2cs}, we compute losses independently for each angle in training, allowing for more nuanced model tuning. 

\textbf{Facial emotion recognition}: We use the unified representation from our backbone as input to a FC layer, which outputs one of eight  emotions. Compared to the resource-intensive transformer models commonly employed in recent emotion recognition research, our approach is more efficient.

\subsection{Three-stage multitask training}
\label{sec:f3}

We developed a structured, three-stage training method across all four tasks to maximize performance and ensure stable feature extraction. In the \textit{first stage}, we train the landmark detection module independently to accurately extract facial landmark information. Once trained, the landmark detection module's parameters are frozen to retain the extracted spatial information throughout the subsequent training stages. In the \textit{second stage}, we focus on stabilizing the classifier features by training only the task-specific classifiers with a frozen backbone. Finally, in the \textit{third stage}, we unfreeze the backbone and task-specific classifiers to fine-tune the interactions between the backbone and the task-specific layers while keeping the pre-trained landmark detection module unchanged. This strategy ensures that the landmark information remains robust and unaffected while allowing the model to optimize its overall performance across tasks.

As we introduce more tasks during training, the complexity increases due to imbalances in data size across different tasks and the fundamentally different nature of loss functions across tasks. To effectively balance these diverse loss functions, we employ a homoscedastic uncertainty-based weighting method \cite{kendall2018multi}. Homoscedastic uncertainty is a type of uncertainty that remains constant across all input data but differs among tasks. This method adjusts the weight of each task’s loss function based on task-dependent uncertainty. Formally, our method assumes that the output of each task with homoscedastic uncertainty follows a Gaussian distribution as described below: 

$$p(\mathbf{y}_{task} \mid f^{\mathbf{W}}(\mathbf{x})) = \mathcal{N}(f^{\mathbf{W}}(\mathbf{x}), \sigma_{task}^2)
$$ where $\mathbf{x}$ is the model input, $W$ is the model weights, $\mathbf{y}_{task}$ is the model output for a specific task, and $\sigma_{task}^2$ is the variance of the specific task. The loss is given by: $$ \begin{aligned}
    &L(W, \hat{\sigma}_{\text{au}}^2, \hat{\sigma}_{\text{gaze}}^2, \hat{\sigma}_{\text{emotion}}^2) = \\
    &\frac{1}{2\hat{\sigma}_{\text{au}}^2} L_{\text{AU}}(W) + \frac{1}{2\hat{\sigma}_{\text{gaze}}^2} L_{\text{gaze}}(W) + \frac{1}{2\hat{\sigma}_{\text{emotion}}^2} L_{\text{emotion}}(W) \\
    &+ \log(\hat{\sigma}_{\text{AU}}^2) + \log(\hat{\sigma}_{\text{gaze}}^2) + \log(\hat{\sigma}_{\text{emotion}}^2)
\end{aligned}
$$
Where $\hat{\sigma}_{task}$ is learnable task dependent parameter.

By maximizing the Gaussian likelihood with homoscedastic uncertainty, this method allows for more principled learning across tasks with different units and scales. 

\subsection{Training Objectives}
\label{sec:f4}
In this section, we discuss the loss functions for each downstream task and \name\ training details.

For \textbf{facial landmark detection}, we use the STAR loss \cite{zhou2023starlossreducingsemantic}, which reduces semantic ambiguity by decomposing prediction error into two principal component directions:

\begin{equation*}
L_{STAR}(y_t, \mu, d) = \frac{1}{\sqrt{\lambda_1}} d(v_1^T (y_t - \mu)) + \frac{1}{\sqrt{\lambda_2}} d(v_2^T (y_t - \mu))
\end{equation*}
where \(y_t\) represents the ground truth coordinates (manual annotations), and \(\mu\) denotes the predicted landmark coordinates obtained through a soft-argmax operation. The distance function \(d(\cdot)\) can be any standard distance measure, such as L1-distance or smooth-L1 distance. The terms \(v_1^T\) and \(v_2^T\) project the prediction error onto the two principal component directions of ambiguity, while \(\lambda_1\) and \(\lambda_2\) are adaptive scaling factors corresponding to the energy of the principal components. By decomposing the error in this manner, the STAR loss helps mitigate ambiguity in facial landmark detection.

 For \textbf{action unit detection}, we use the Weighted Asymmetric Loss \cite{Luo_2022}, which handles class imbalance by assigning higher weights to less frequent AUs:
\begin{equation*}
L_{WA} = - \frac{1}{N} \sum_{i=1}^{N} w_i [y_i \log(p_i) + (1 - y_i) p_i \log(1 - p_i)]
\end{equation*}
where \(w_i\) is the weight assigned to the \(i\)-th action unit, inversely proportional to its occurrence in the training set. The term \(y_i\) is the ground truth label for the \(i\)-th action unit, and \(p_i\) is the predicted probability of its occurrence. The term \((1 - y_i)p_i \log(1 - p_i)\) down-weights the loss for inactivated AUs that are easier to predict, thus focusing learning on more challenging AUs. This weighting mechanism ensures that both frequent and rare AUs are balanced during training.

For \textbf{gaze estimation}, we apply the Mean Squared Error (MSE) loss, which minimizes the squared difference between the predicted and true gaze angles.

For \textbf{facial emotion recognition}, we use a class-weighted cross-entropy loss with label smoothing \cite{savchenko2022classifying} to address the common issue of class imbalance.  Given a class label \( y \in \{1, \dots, C\} \), we first apply label smoothing to obtain a soft target distribution:

\begin{equation*}
\tilde{y}_i = (1 - \alpha) \delta_{i,y} + \frac{\alpha}{C}, \quad i \in \{1, \dots, C\}
\end{equation*}where \( \alpha \) is the label smoothing factor, \( C \) is the total number of classes, and \( \delta_{i,y} = 1 \) if \( i = y \) and 0 otherwise. The final loss function is defined as:

\begin{equation*}
\mathcal{L} = - \frac{1}{N} \sum_{i=1}^{N} \sum_{j=1}^{C} \frac{f_j}{\min f_k } \tilde{y}_{i,j} \log p_{i,j}
\end{equation*}
where \( f_j \) is the inverse frequency of class \( j \) in the dataset, and \( p_{i,j} \) is the predicted probability for class \( j \) of sample \( i \).

\name \ is trained PyTorch using the AdamW optimizer with momentum  $\beta_1 = 0.9$ and $\beta_2 = 0.999$. A weight decay of $5 \times 10^4$ is applied to prevent overfitting. For the three training stages (Section \label{sec:f3}), we use different learning rates and epoch numbers for different training goals. In the first stage, the landmark detection module is trained for 100 epochs with a learning rate of 0.001. In the second stage, parameters of the shared backbone are frozen. The training focuses on aligning initialized task-specific layers with the unified features extracted by the pre-trained multitasking backbone. As only a limited number of parameters are updated, the model is trained for 5 epochs with a learning rate of  $0.001$.  In the final stage, the training aims to finetune the entire model. This stage employs a smaller learning rate of $0.0001$ and a larger number of epochs ($15$) to ensure fine-grained optimization across all tasks. All training for \name\ was done with a single GTX1080Ti GPU.

\subsection{\name\ Interface and Deployment}
\label{f5}

\name\ offers a versatile, well-documented, open-source toolkit for researchers and developers interested in analyzing facial behavior in their own datasets. Expertise in computer vision or machine learning is not necessary for researchers to rapidly leverage \name\ in their projects. 
\name\ is capable of processing real-time video feeds from webcams, recorded video files, image sequences, and individual images. The system’s inference speed has been optimized, outperforming off-the-shelf facial analysis tools and ensuring that \name\ can operate efficiently in real-time use cases such as social signal processing, human-robot interaction, and multimodal interfaces. 

\name\ is implemented in PyTorch. We introduce a new Python package that can be easily installed via the \texttt{pip} command. \name\ also includes a command line interface for added flexibility. To help users quickly deploy and utilize the toolkit, we provide sample Python scripts to demonstrate how to extract, save, read, and visualize facial behavior predictions, making it seamless for users to integrate \name\ into their workflows. 



%

\begin{table*}[t]
\fontsize{8}{9}\selectfont
\setlength\tabcolsep{2.0pt}
\centering
\caption{Summary of all datasets standardized and used to train \name.}
\vspace{-2mm}
\begin{tabular}{lccl}
\toprule
Datasets & Size & Task & Description \\
\midrule
300W \cite{sagonas2016300} & 4,437 images & Landmark Detection & 68 manually annotated landmarks per image \\
WFLW \cite{wayne2018lab} & 10,000 images & Landmark Detection & 98 landmarks across diverse faces \\
AffectNet \cite{mollahosseini2017affectnet} & 1,000,000 images & Emotion Recognition & 8 categorical emotions with valence and arousal scores \\
DISFA \cite{mavadati2013disfa} & 27 hours video & Action Unit Recognition & Intensity of 12 action units on a 6-point scale \\
BP4D \cite{zhang2014bp4d} & 192,000 frames of video & Action Unit Recognition & 27 action units in 2D and 3D expressions \\
MPII Gaze \cite{zhang2015appearance} & 213,659 images & Gaze Estimation & Gaze estimation in natural environments \\
Gaze360 \cite{kellnhofer2019gaze360} & 200,000 images & Gaze Estimation& Gaze directions covering a full 360-degree range \\
\bottomrule
\end{tabular}
\vspace{1mm}
\label{tab:all_datasets}
\end{table*}

\begin{table*}[t]
\fontsize{8}{9}\selectfont
\setlength\tabcolsep{2.0pt}
\centering
\caption{Comparison of state-of-the-art models across tasks.  }
\begin{tabular}{lccl}
\toprule
Model & Task & Parameters(M) & Description \\
\midrule
SPIGA \cite{prados2022shape} & Landmark Detection & 63.3  & Combines CNNs with Graph Attention Networks to model spatial relationships. \\
SLPT \cite{xia2022sparse} & Landmark Detection & 13.2 & Lightweight model using transformers for subpixel coordinate prediction. \\
ME-GraphAU \cite{Luo_2022} & Action Unit Recognition & 67.4   & Graph-based approach with dynamic edge features to model AU dependencies. \\
LibreFace \cite{Chang_2024_WACV} & Action Unit Recognition &  22.5 & Toolkit for real-time AU detection with feature-wise knowledge distillation. \\
MCGaze \cite{guan2023end} & Gaze Estimation &  83.1 & Models spatial-temporal interactions for robust gaze direction estimation. \\
L2CS-Net \cite{abdelrahman2023l2cs} & Gaze Estimation &  41.6  & Multi-loss, two-branch CNN architecture for angle-specific gaze prediction. \\
DDAMFN \cite{zhang2023dual} & Emotion Recognition & 4.1  & Uses dual-direction attention mechanism for global and local feature integration. \\
EfficientFace \cite{zhao2021robust}  & Emotion Recognition & 1.3 & Lightweight model using depthwise convolutions and label distribution learning. \\
Py-Feat \cite{cheong2023py} & Landmark, AU, Emotion & 49.3 & A versatile facial analysis toolkit integrating multiple deep learning models. \\
OpenFace 2.0 \cite{8373812} & Landmark, AU, Gaze & 44.8  & A popular open-source facial analysis toolkit. \\
\name\  & Landmark, AU, Gaze, Emotion & 29.4  & Unified multitask model with an efficient backbone. \\
\bottomrule
\end{tabular}
\vspace{-2mm}
\label{tab:model_comparison}
\end{table*}

\section{Experiments}
Our experiments are designed to test the performance and efficiency of using \name. We begin with quantitative experiments to benchmark various performance metrics of \name. All experiments discussed in this section were conducted on a machine equipped with an AMD Ryzen Threadripper 1920X @ 3.5GHz CPU and an NVIDIA GeForce GTX 1080 Ti GPU.

\subsection{Data and baselines}
 In this section, we describe datasets and metrics used to evaluate performance on each of the facial analysis tasks included in \name. For each task, we list SOTA baselines to which we compare, as well as lightweight (smaller) baseline models that we designate as \textit{SOTA small models}. We include lightweight model baselines to assess the feasibility of using \name\ in resource-constrained environments where speed and efficiency are critical (e.g., real-time edge computing, robotics). We also conduct experiments for each task with the widely-used \textsc{OpenFace 2.0} toolkit \cite{8373812} as a baseline for landmark detection, AU detection, and gaze estimation (emotion recognition is not supported by \textsc{OpenFace 2.0}). We conduct experiments, when applicable, with Py-Feat \cite{cheong2023py} and LibreFace \cite{Chang_2024_WACV} -- these toolkits cannot perform all the downstream tasks. All datasets and models discussed below are summarized in Table \ref{tab:all_datasets} and Table \ref{tab:model_comparison}. 

\subsubsection{\textbf{Facial landmark detection}}




\textit{Datasets:} The 300W dataset \cite{sagonas2016300} includes 68 landmarks per image and is divided into indoor and outdoor conditions, supporting robust training and evaluation. The WFLW dataset \cite{wayne2018lab} provides a diverse array of faces with 98 landmarks, covering variations in occlusion, pose, make-up, illumination, blur and expression.

\textit{Baselines:} SPIGA \cite{prados2022shape}, a SOTA model, combines CNNs with Graph Attention Networks (GAT) to capture  local appearance and spatial relationships among landmarks, enhancing detection accuracy. SLPT \cite{xia2022sparse} is a SOTA lightweight model that leverages a local patch transformer to learn relationships among landmarks, generating representations from small local patches and using attention mechanisms to predict landmark subpixel coordinates. We report Py-Feat performance for the 300W dataset; Py-Feat's landmark detection cannot perform the WFLW task out-of-the-box. LibreFace's face alignment outputs a 5-point landmark set and has not been trained to detect 68 or 98 facial landmarks. 

\textit{Metrics:} Normalized Mean Error (NME\_int\_ocul) is used to evaluate landmark detection accuracy. NME measures the average Euclidean distance between the predicted and ground-truth landmarks, normalized by the inter-ocular distance (distance between the outer eye corners). 

\subsubsection{\textbf{Facial action unit recognition}}



\textit{Datasets:} DISFA \cite{mavadati2013disfa} contains videos annotated for the intensity of 12 common facial AUs, allowing detailed analysis of facial dynamics. BP4D \cite{zhang2014bp4d} includes both 2D and 3D facial expressions annotated for 27 action units, covering diverse demographics and supporting comprehensive facial expression analysis.

\textit{Baselines:} ME-GraphAU \cite{Luo_2022} is a SOTA model that constructs an AU relation graph with multi-dimensional edge features to capture dependencies between action units, enhancing recognition accuracy and robustness. LibreFace \cite{Chang_2024_WACV} is an open-source facial analysis tool with SOTA lightweight models for real-time AU detection, utilizing feature-wise knowledge distillation. 

\textit{Metrics:} The F1 score is used to evaluate the performance of AU detection, providing a harmonic mean that balances false positives and false negatives. 

\begin{table*}[t]
    \fontsize{8}{9}\selectfont
    \setlength\tabcolsep{2.0pt}
    \centering
    \caption{\name \ represents our model trained independently on each task. The highest-performing model results are \textbf{bolded}, and \name\ results that outperform SOTA small models are highlighted in \setlength{\fboxsep}{1pt} \colorbox{lightblue}{blue}. \name \ (MTL) refers to our model trained with multitask learning. \name \ (MTL w/ unc.) refers to our model trained with uncertainty loss during multitask learning. Across all tasks, our approach either exceeds or performs comparably to SOTA models, SOTA (small models), and other toolkits trained for specialized tasks. The '-' symbol indicates the model does not perform the task.}
    \vspace{-0mm}
    \begin{tabular}{lcccccccc}
    \midrule
    Models & \multicolumn{2}{c}{Landmark Detection} & \multicolumn{2}{c}{Gaze Estimation} \\
    \cmidrule(lr){2-3} \cmidrule(lr){4-5}
    & 300W (NME\_int\_ocul $\downarrow$) & WFLW (NME\_int\_ocul $\downarrow$) & MPII (diff. in angle $\downarrow$) & Gaze360 (diff. in angle $\downarrow$) \\
    \midrule
    OpenFace 2.0 & 5.20 & 7.11 & 9.10 & 26.7 \\
    SOTA &  2.99 & \textbf{4.00} & 3.14 & \textbf{10.0} \\
    SOTA (small models) & 3.17 & 4.14 & 3.92 & 10.4 \\
    LibreFace & - & - & - & - \\
    Py-Feat & 4.99 & - & - & -\\
    \name\ & 2.87 & 4.02 & 4.62 & 13.7 \\
    \name\ (MTL) & 2.87 & 4.02 & 4.25 & 10.7 \\
    \name\ (MTL w/ unc.) & \fcolorbox{white}{lightblue}{\textbf{2.87}}  & \fcolorbox{white}{lightblue}{4.02} & \fcolorbox{white}{lightblue}{\textbf{2.56}} & 10.6 \\
    \midrule
    Models & \multicolumn{2}{c}{Action Unit Detection} & \multicolumn{2}{c}{Facial Emotion Recognition} \\
    \cmidrule(lr){2-3} \cmidrule(lr){4-5}
    & DISFA (F1 Score $\uparrow$) & BP4D (F1 Score $\uparrow$) & \multicolumn{2}{c}{AffectNet (8 emotion ACC $\uparrow$)} \\
    \midrule
    OpenFace 2.0 & 50 & 53 & \multicolumn{2}{c}{-} \\
    SOTA & \textbf{66} & \textbf{66} & \multicolumn{2}{c}{\textbf{0.65}} \\
    SOTA (small models) & 61 & 62 & \multicolumn{2}{c}{0.60} \\
    LibreFace  & 61 & 62 & \multicolumn{2}{c}{0.49} \\
    Py-Feat & 54 & 55 & \multicolumn{2}{c}{-}\\
    \name\ & 56 & 57 & \multicolumn{2}{c}{0.59} \\
    \name\ (MTL) & 60 & \fcolorbox{white}{lightblue}{62} & \multicolumn{2}{c}{0.56} \\
    \name\ (MTL w/ unc.) & 59 & 59 & \multicolumn{2}{c}{\fcolorbox{white}{lightblue}{0.60}} \\
    \bottomrule
    \end{tabular}
    \vspace{1mm}
    
    \vspace{-2mm}
    \label{tab:human_eval}
\end{table*}

\subsubsection{\textbf{Eye gaze estimation}}



\textit{Datasets:} MPII Gaze \cite{zhang2015appearance} contains gaze data collected in natural environments, capturing variability in appearance and illumination from participants using laptops in everyday settings. Gaze360 \cite{kellnhofer2019gaze360} includes annotated gaze directions spanning a full 360-degree range, covering a wide array of head poses and lighting conditions, ideal for training models for dynamic real-world scenarios.

\textit{Baselines:} MCGaze \cite{guan2023end} is a SOTA model that enhances gaze estimation accuracy by capturing spatial-temporal interactions between the head, face, and eyes. L2CS-Net \cite{abdelrahman2023l2cs} is a lightweight SOTA model that uses a multi-loss, two-branch CNN architecture to predict each gaze angle with separate fully connected layers, improving estimation precision. Both LibreFace and Py-Feat do not offer gaze estimation.

\textit{Metrics:} Difference in angles (dff. in angle) is used to evaluate gaze estimation, computing angular distance between predicted and ground-truth gaze directions in 3D space.

\subsubsection{\textbf{Facial emotion recognition}}



\textit{Datasets:} AffectNet \cite{mollahosseini2017affectnet} includes a wide range of expressions, with labels for eight categorical emotions—neutral, happiness, sadness, surprise, fear, disgust, anger, and contempt. This  dataset is valuable for training robust facial expression recognition systems.

\textit{Baselines:} DDAMFN \cite{zhang2023dual} is a SOTA model that uses a dual-direction attention mechanism to combine global and local features, enhancing expression recognition accuracy. EfficientFace \cite{zhao2021robust} is a lightweight SOTA model that employs a local-feature extractor and a channel-spatial modulator with depthwise convolution to capture both local and global features, along with label distribution learning to handle complex emotion combinations. Both \textsc{Openface 2.0} and Py-Feat do not provide emotion recognition function.

\textit{Metrics:} Accuracy is used to evaluate the performance of facial expression recognition. 

\subsection{Results}

Results from experiments with \name\ and baseline models are summarized in Table~\ref{tab:human_eval}. For each task and dataset, the highest-performing model results are bolded, and \name\ results that outperform SOTA small models have a blue highlight. \textit{Overall, \name\ achieves strong performance across all 4 tasks, either exceeding or performing comparably to SOTA models or SOTA (small models) that were trained for specialized tasks. In addition, \name\ substantially outperforms \textsc{OpenFace 2.0} across all tasks.} These findings indicate that \name\ contributes a useful  toolkit for researchers and developers performing multiple facial analysis tasks.   

\textbf{Facial landmark detection results}: Our model achieves NME of 2.87 on the 300W dataset, a substantial improvement when compared to both \textsc{OpenFace 2.0} (5.20) and SOTA small models (3.17), and comparable to SOTA results (2.99). For the WFLW dataset, our model achieves an NME of 4.02, a substantial improvements from \textsc{OpenFace 2.0} (7.11), outperforming SOTA small models (4.14), and comparable to SOTA results on this task (4.00). 

\textbf{AU detection results}: Our model achieves F1 scores of 60\% and 62\%, respectively on the DISFA and BP4D datasets, substantially outperforming the \textsc{OpenFace 2.0} performance (50\% and 53\%, respectively) and achieving comparable performance to SOTA small models (61\% and 62\%, respectively). We note that our comparable performance to smaller SOTA models for AU detection indicates that \name\ can be useful in facial analysis contexts requiring smaller models to perform diverse tasks. Unlike  specialized SOTA smaller AU detection models, \name\ can also perform landmark detection, gaze estimation, and facial emotion recognition when deployed in contexts with less compute (e.g., edge devices). 

\textbf{Gaze estimation results}: Our model achieves an angular error of 2.56 on MPII Gaze, outperforming both \textsc{OpenFace 2.0} (9.1) and SOTA (3.14). Although performance gains are less pronounced on the more challenging Gaze360 dataset, our model (10.6) achieves comparable results to both SOTA (10.0) and SOTA small models (10.4) that were trained specifically for this tak, demonstrating our model's strong performance in this integrated toolkit alongside other tasks. 

\textbf{Emotion recognition results}: Our model attains a competitive accuracy of 0.60 on AffectNet (compared to 0.65 SOTA and 0.60 SOTA small models) without employing advanced transformer-based techniques, demonstrating the strong performance of our toolkit's approach. We also find that our model outperforms both the SOTA and SOTA small model for emotion recognition when performing inferences on non-frontal faces (additional discussion in Section \ref{subsec:ablation}), a key capability for toolkits operating on data in-the-wild. 

\textbf{Efficiency Comparisons}: We anticipate that \name\ will be most useful to researchers and developers building real-time facial analysis applications in domains such as multimodal data analysis, human-machine interaction, and robotics. Therefore, model efficiency while performing these 4 downstream tasks is of critical importance. Table \ref{tab:model_efficiency} presents the comparison between \name\ and other widely-used facial toolkits (\textsc{OpenFace 2.0}, LibreFace, Py-Feat) in terms of CPU-only processing speed. By employing a multitasking model with shared features, \name\ contains fewer parameters (29.4 million) than \textsc{OpenFace 2.0} (44.4 million) and Py-Feat (49.3 million), and demonstrates a faster processing speed that better supports real-time applications. Although LibreFace is slightly faster, \name\ performs a broader range of tasks than Libreface while maintaining a similar speed -- the Libreface toolkit supports AU detection and emotion recognition, whereas \name\ performs all 4 facial analysis tasks. 

\begin{figure*}[ht]
    \centering
    \includegraphics[width=0.85\textwidth]{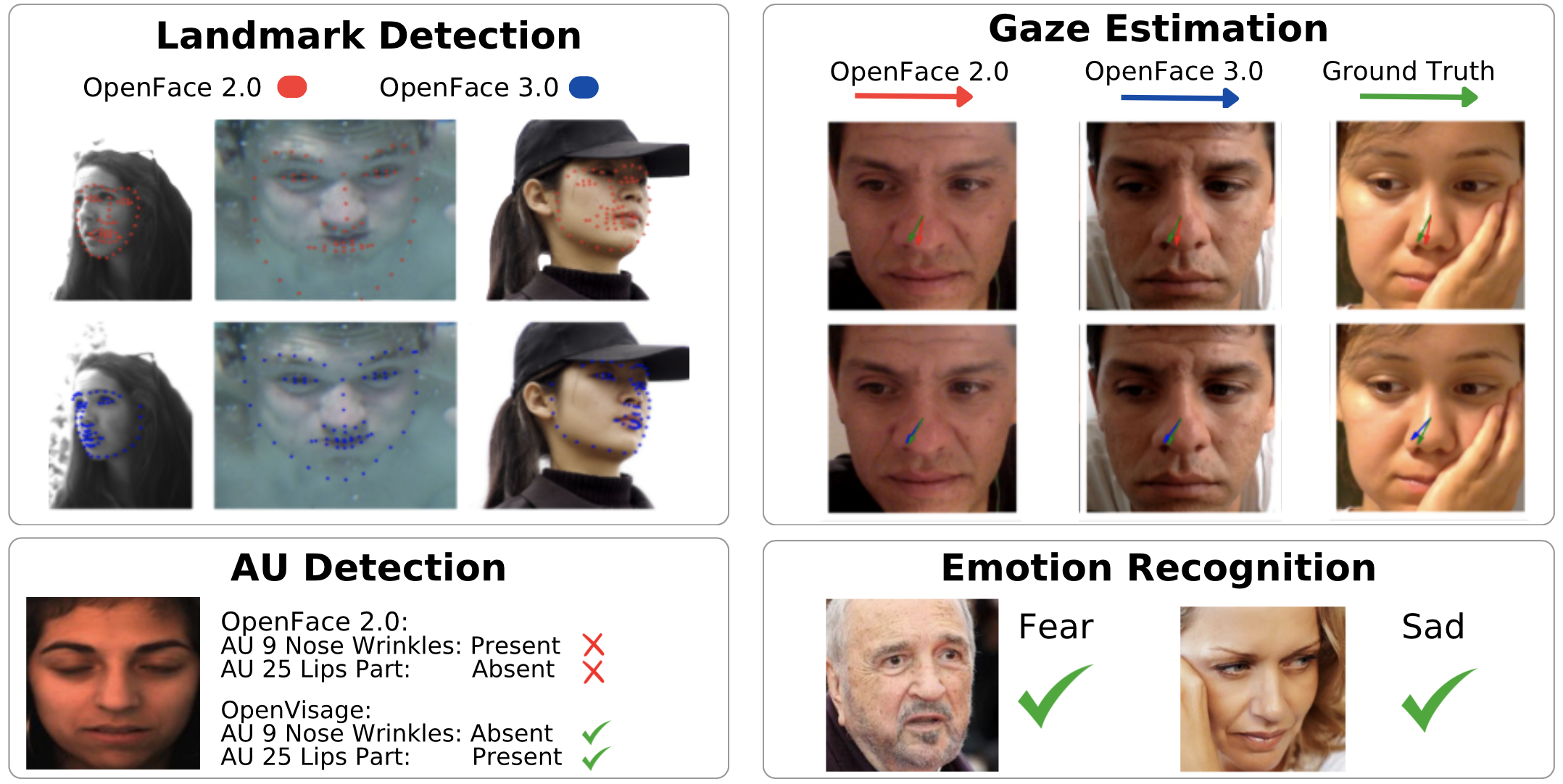}
    \caption{Samples of task performance between \name\ and \textsc{OpenFace 2.0}. \textsc{OpenFace 2.0} does not support emotion recognition.}
    \label{fig:all_comp}
\end{figure*}

To qualitatively demonstrate our model's notable improvements, we visualize the performance of \name\ and \textsc{OpenFace 2.0} on several examples in Figure \ref{fig:all_comp}. We select \textsc{OpenFace 2.0} for comparison as it is the toolkit that provides the most similar set of functionalities. These examples demonstrate that \name\ predictions, in particular for non-frontal faces, are more accurate than predictions from \textsc{OpenFace 2.0}. Overall, our model balances performance, efficiency, and adaptability, making it well-suited for real-time, multi-task facial analysis applications.

\subsection{Ablation Studies}
\label{subsec:ablation}
In this section, we evaluate the significance of the techniques used in our model, particularly the impact of multitasking and uncertainty weighting. We conduct an ablation study across three downstream tasks that utilize the shared backbone: AU detection, emotion recognition, and gaze estimation. Our goal is to determine how multitasking influences the performance on each task and the interplay between these tasks when using a shared backbone.


\textbf{Multi-task learning and uncertainty weighting}: We hypothesize that multitasking with uncertainty weighting improves generalization by sharing learned features across tasks, benefiting tasks with overlapping data characteristics while balancing task-specific requirements. To assess this, we examine the performance of the single-task model (\name), multitask model (MTL), and multitask model with uncertainty weighting (MTL w/ unc.) in Table~\ref{tab:human_eval}.

Introducing multitasking (MTL) substantially improves AU detection, with the F1 score increasing from 56\% to 60\% on the DISFA dataset and from 57\% to 62\% on BP4D. The gaze estimation with the MTL model also improved from 4.62 angular error to 4.25 angular error on MPII and from 13.7 angular error on Gaze360 to 10.7 angular error. 

However, multitasking slightly lowers emotion recognition accuracy (from 0.59 to 0.56 on AffectNet). This suggests that the shared feature in the multitasking model benefits tasks by offering additional information learned from various tasks. The multitasking setup also introduces trade-offs for tasks, likely due to data imbalance and the varying loss functions.

Uncertainty weighting addresses these trade-offs by introducing task-specific weights, balancing each task’s contribution. With uncertainty weighting (MTL w/ unc.), AU detection performance remains stable around an F1 score of 0.59, while gaze estimation achieved further improvement, substantially reducing the angular error from 4.25 in the multitasking model to 2.56 on MPII and from 10.7 on Gaze360 to 10.6. For emotion recognition, accuracy with the uncertainty weighting model \textit{increases} slightly to 0.60, surpassing the single-task baseline. These results suggest that multitasking benefits tasks more fully once data and loss imbalances are resolved with uncertainty weighting.

\begin{table}[t]
    \fontsize{8}{9}\selectfont
    \setlength\tabcolsep{2.0pt}
    \centering
    \caption{Model efficiency with CPU-only processing.}
    \vspace{-0mm}
    \begin{tabular}{lccc}
    \toprule
    Models & \# Params  & \# Operation & Processing Time\\
     &  (M)  & (GFLOPs) & Per Frame (ms) \\
    \midrule
    OpenFace 2.0 & 44.8 & 7.4 & 75 \\
    LibreFace & 22.5 & 3.7 & 33 \\
    Py-Feat & 49.3 & 8.2 & 82\\
    \name\ & 29.4 & 4.2 & 38 \\
    \bottomrule
    \end{tabular}
    \vspace{1mm}
    \vspace{-2mm}
    \label{tab:model_efficiency}
\end{table}

\begin{table}[t]
    \fontsize{8}{9}\selectfont
    \caption{Performance of DDAMFN (SOTA), EfficientFace (SOTA small) and \name\ for emotion recognition across samples grouped by face orientation.}
    \begin{tabular}{lccc}
    \toprule
    \textbf{Orientation}  & \textbf{DDAMFN} & \textbf{EfficientFace} & \textbf{\name} \\
    \midrule
    Easy (0°–15°) & 62.74 & 58.90 & 60.20 \\
    Medium (15°–45°) & 61.99 & 55.19 & 58.21 \\
    Hard ($>$ 45°) & 57.41 & 53.97 & \textbf{58.64} \\
    \bottomrule
    \end{tabular}
    \label{tab:orientation_performance}
\end{table}

\textbf{Angled Faces Analysis}:  We analyze the model’s performance on non-frontal (angled) faces, a common challenge in most in-the-wild facial analysis tasks. In our multitasking model, we hypothesize that shared features learned from tasks with non-frontal training data—specifically, gaze estimation—will improve performance across all tasks for non-frontal faces, especially emotion recognition which only has a small number of non frontal faces in training data.

To evaluate this, we divided the AffectNet validation set based on face orientation into three categories: Easy (0°–15°), Medium (15°–45°), and Hard ($>$45°). This setup allowed us to assess if features learned from non-frontal gaze data enhance robustness across orientations in other tasks. As seen in Table~\ref{tab:orientation_performance}, for a specialized SOTA model DDAMFN \cite{zhang2023dual} and lightweight model EfficientFace \cite{zhao2021robust} that are only trained on emotion data, the accuracy declines as orientation becomes more angled. Notably, our multitasking model shows stable performance across all angles compared to the single-task baselines. \name \ even exceeds the SOTA model on the hard set. This indicates that features shared from other tasks contribute to handling non-frontal faces across tasks even without explicit emotion labels.

These results demonstrate that our multitasking model’s adaptability can benefit performance in real-world settings with varied facial orientations. Future research might explore techniques for incorporating more non-frontal data to further boost performance on angled faces.

\section{User Study}

We conducted a pilot user study to evaluate the usability of \name, focusing on the accessibility for users with varying background familiarity with computer vision. In our study, we recruited 8 graduate students with different familiarity levels with machine learning or computer vision (beginner, intermediate, advanced) to set up and test \name\ from scratch (details in Appendix \ref{sec:appendix}). Participants were asked to perform several standard tasks with our system, divided into installation and setup, image-based, and video-based evaluations:

    


\begin{itemize}
    \item \textbf{Installation and Setup}: Install \name\ using our  guide and run a demo program to confirm the setup.

    \item \textbf{Image-Based Tests}: 1) Extract and overlay left eye gaze direction. 2) Identify primary emotion in an image. 3) Measure intensity of AU 26 (mouth drop).
    
    \item \textbf{Video-Based Tests}: 1) Track gaze direction with frame-by-frame overlays. 2) Identify most common emotions across frames. 3) Measure average intensity of AU 12 (lip corner puller) over time.
    
\end{itemize}
\raggedbottom

Participants completed the UMUX-LITE \cite{lewis2013umux} and NASA-TLX \cite{hart2006nasa} questionnaires after  each task. Each questionnaire provides insights into the participant experience, with UMUX-LITE capturing general usability and NASA-TLX evaluating perceived workload across multiple dimensions. The UMUX-LITE scores are based on a 7-point scale, with lower scores indicating higher usability. \textit{Usability} refers to participants' overall satisfaction with the tool’s ability to meet their requirements, while \textit{Ease of Learning} assesses participants' perceived ease of learning how to use \name. The NASA-TLX scores are on a 20-point scale, where lower scores reflect lower perceived workload.


Appendix Table~\ref{tab:questionnaire} summarizes the usability scores for each task set in our pilot study. These scores demonstrate that installation, overall, was perceived as the easiest task, as indicated by both low UMUX-LITE scores and the lowest NASA-TLX workload demands. Image-based and video-based tasks, while slightly more challenging, still maintained relatively low workload and usability scores, suggesting that \name\ can function as a user-friendly system, even for participants with varying levels of experience.




\vspace{-1mm}
\section{Conclusion}
\vspace{-1mm}

Processing communicative signals from human faces will be a critical component of future AI systems deployed in-the-wild to reason about human behavior and interact with humans. These  systems will need to be \textit{lightweight}, \textit{accurate}, and capable of performing multiple downstream facial analysis tasks, in order to operate effectively on both static behavior datasets and  real-time human-machine interactions. In this paper, we introduce \name, a new open-source, lightweight, multi-task system for facial behavior analysis, capable of jointly performing key downstream tasks: landmark detection, action unit detection, eye gaze estimation, and emotion recognition. \name\ leverages a unified multitask learning approach trained on diverse populations, poses, lighting conditions, and video resolutions, making it robust and adaptable to in-the-wild contexts. In comparison to prior facial analysis toolkits and SOTA models for specialized downstream tasks, \name\ performs a wider variety of tasks with improvements in task performance, efficiency, inference speed, and memory usage. Additionally, \name\ runs in real-time on standard hardware, requiring no specialized equipment. 

Beyond the technical advantages of \name, this system is available as a fully \textit{open-source} toolkit, encouraging transparency in facial analysis research and community contributions and use in multimodal interaction, robotics, and affective computing. Our system is easy to install for users, is built upon an intuitive Python API, and can be deployed with a single line of code to extract diverse facial cues in real-time from images and video streams. We believe that the accessibility, efficiency, and strong performance of \name\ will stimulate continued advancements in automatic facial behavior analysis.

\section{Ethical Impact Statement}
The data used by \name\ does not involve direct interaction with human participants or the study of private or confidential human data. Our research utilizes publicly-available datasets to train models for for facial analysis tasks. Our project includes a pilot user study, in which graduate students in our community tested the \name\ installation process and completed demo tasks; this pilot study did not require an IRB review. All surveys conducted as part of the user study are anonymous, with no personally-identifiable information collected. 

\paragraph{Potential Risks} The primary risks associated with \name\ stem from potential misuse of facial analysis technologies in downstream applications. Computer vision systems to predict facial landmarks, action units, eye gaze, and emotion have great potential to support humans -- for example, gaze tracking can improve the effectiveness of interactive interfaces and social robot companions for the elderly. However, broader risks associated with automated facial analysis systems include potential biases in deployed models and possible misuse in surveillance applications. 

\paragraph{Benefit-Risk Analysis} \name\ aims to advance research in facial analysis, while prioritizing ethical considerations. The project provides a valuable, open-source tool for researchers, developers, and educators working on applications that require facial analysis. The pilot user study helped us refine the installation and usability of the software without posing risks to any participants. \name\ facilitates facial analysis research, enables transparent development of this research, and provides a more ethical alternative to proprietary facial analysis systems. 

\section*{Acknowledgments}
This material is based upon
work partially supported by National Institutes of
Health awards R01MH125740, R01MH132225,
and R21MH130767. Leena Mathur is supported by the NSF Graduate
Research Fellowship Program under Grant
No. DGE2140739. Any opinions, findings, conclusions,
or recommendations expressed in this material
are those of the authors, do not necessarily
reflect the views of any sponsors, and no official
endorsement should be inferred. 

{\small
\bibliographystyle{ieee}
\bibliography{main}
}

\ \

\section*{Appendix}
\label{sec:appendix}

For our pilot study of \name\ usability, we tested the system with 8 graduate students with diverse levels of prior experience with computer vision and machine learning. Participants were classified based their prior experience:
\begin{itemize}
\item \textbf{Novice (N = 3)}: Participants had minimal or no prior experience with computer vision tools or machine learning frameworks.
\item \textbf{Intermediate (N = 3)}: Participants had some familiarity, having used computer vision or machine learning tools occasionally for coursework or personal projects.
\item \textbf{Experienced (N = 2)}: Participants regularly used computer vision and machine learning tools in their research or professional contexts.
\end{itemize}

Appendix Table \ref{tab:questionnaire} contains usability scores in our pilot study across each of the 3 tasks that users performed: installation and setup, image-based feature extraction and visualization, and video-based feature extraction and visualization. The UMUX-LITE \cite{lewis2013umux} scale studies general \textit{usability} of systems (does the system meet a user's requirements and how easy is it to use?). The NASA-TLX \cite{hart2006nasa} studies \textit{perceived workload} across dimensions, including mental demand (mental activity required), physical demand (physical activity/strain required), temporal demand (time pressure felt while using the system), performance (success and satisfaction in performing a task), effort, and frustration. 

Appendix Table~\ref{tab:experience_breakdown} presents the breakdown of UMUX-LITE usability and NASA-TLX workload scores by participant experience levels. Novice participants, on average, reported higher workload and lower usability compared to experienced participants, particularly in video-based tasks. This indicates that while \name\ remains accessible to researchers and developers across varying experience levels, novices may require additional guidance or more detailed Python package documentation. 

\begin{table}[h]
    \centering
    \fontsize{8}{9}\selectfont
    \caption{\name\ usability scores for each task set.}
    \begin{tabular}{lccc}
    \toprule
    & \textbf{Install} & \textbf{Image Tasks} & \textbf{Video Tasks} \\
    \midrule
    \textbf{UMUX-LITE (1-7)} & & & \\
    Usability & 2 & 3 & 3 \\
    Ease of Learning & 2 & 3 & 3 \\
    \midrule
    \textbf{NASA-TLX (1-20)} & & & \\
    Mental Demand & 5 & 10 & 12 \\
    Physical Demand & 3 & 4 & 5 \\
    Temporal Demand & 4 & 8 & 10 \\
    Performance & 4 & 7 & 8 \\
    Effort & 5 & 9 & 10 \\
    Frustration & 3 & 5 & 6 \\
    \bottomrule
    \end{tabular}
    \label{tab:questionnaire}
\end{table}

\begin{table}[h]
\centering
\fontsize{8}{9}\selectfont
\caption{\name\ usability scores by participant experience levels.}
\begin{tabular}{lccc}
\toprule
& \textbf{Novice} & \textbf{Intermediate} & \textbf{Experienced} \\
\midrule
\textbf{UMUX-LITE (1-7)} & & & \\
Usability & 4 & 3 & 2 \\
Ease of Learning & 4 & 3 & 2 \\
\midrule
\textbf{NASA-TLX (1-20)} & & & \\
Mental Demand & 14 & 10 & 7 \\
Physical Demand & 5 & 4 & 3 \\
Temporal Demand & 12 & 8 & 5 \\
Performance & 10 & 7 & 5 \\
Effort & 13 & 9 & 6 \\
Frustration & 9 & 5 & 3 \\
\bottomrule
\end{tabular}
\label{tab:experience_breakdown}
\end{table}
\end{document}